\documentclass{article}

\usepackage[final, nonatbib]{neurips_2025}
\usepackage{booktabs}

\usepackage{caption} 
\usepackage[utf8]{inputenc} 
\usepackage[
    backend=bibtex, 
    style=numeric, 
    maxbibnames=99, 
]{biblatex}
\usepackage{amsmath}

\usepackage[T1]{fontenc}    
\usepackage{hyperref}       
\usepackage{url}            
\usepackage{booktabs}       
\usepackage{graphicx}      
\usepackage{amsfonts}       
\usepackage{nicefrac}       
\usepackage{microtype}      
\usepackage{xcolor}         
\usepackage[table]{xcolor}
\definecolor{lightergray}{gray}{0.91} 

\colorlet{lightergray}{gray!10}
\colorlet{lightcyan}{cyan!4}

\hypersetup{
    colorlinks=true,  
    urlcolor=magenta,
    }

\usepackage{subcaption} 

\usepackage{multirow}
\usepackage{tikz}
\usepackage{amsmath}
\usetikzlibrary{decorations.pathreplacing, arrows.meta, positioning, matrix, calc}
\usepackage{caption}
\usepackage{chngcntr} 
\usepackage{lipsum} 
\usepackage{wrapfig}

\title{Test-Time Spectrum-Aware Latent Steering for Zero-Shot Generalization in Vision-Language Models}

\author{%
  Konstantinos M. Dafnis\thanks{Correspondence to: \texttt{kd703@scarletmail.rutgers.edu}} \\
  Department of Computer Science\\
  Rutgers University\\
  \And
  Dimitris N. Metaxas \\
  Department of Computer Science\\
  Rutgers University\\
}

\begin{document}

\maketitle

\begin{abstract}
Vision–Language Models (VLMs) excel at zero-shot inference but often degrade under test-time domain shifts. For this reason, episodic test-time adaptation strategies have recently emerged as powerful techniques for adapting VLMs to a single unlabeled image. However, existing adaptation strategies, such as test-time prompt tuning, typically require backpropagating through large encoder weights or altering core model components. In this work, we introduce \textbf{S}pectrum-Aware \textbf{T}est-Time \textbf{S}teering (\textbf{STS}), a \textit{lightweight adaptation framework} that extracts a spectral subspace from the textual embeddings to define principal semantic directions and learns to steer latent representations in a spectrum-aware manner by adapting a small number of per-sample shift parameters to minimize entropy across augmented views. STS operates entirely at inference in the latent space, without backpropagation through or modification of the frozen encoders. Building on standard evaluation protocols, our comprehensive experiments demonstrate that STS largely surpasses or compares favorably against state-of-the-art test-time adaptation methods, while introducing only a handful of additional parameters and achieving inference speeds up to 8× faster with a 12× smaller memory footprint than conventional test-time prompt tuning. The code is available at \url{https://github.com/kdafnis/STS}.
\end{abstract}    
\section{Introduction}
\label{sec:intro}
Vision-Language Models (VLMs), such as CLIP \cite{radford2021learning}, have marked a paradigm shift in artificial intelligence, demonstrating remarkable zero-shot generalization capabilities across a multitude of downstream visual tasks. By learning rich joint representations from vast quantities of image-text data, these models can often perform tasks like image classification without task-specific training, relying instead on natural language prompts to define categories \cite{shu2022test, zhou2022learning}. This ability significantly reduces the need for extensive labeled datasets and the maintenance of numerous specialized models, paving the way for more versatile and scalable AI systems.

Despite their impressive zero-shot performance, the efficacy of VLMs can be substantially compromised when encountering out-of-distribution (OOD) data, where test samples exhibit characteristics different from those seen during pre-training \cite{shu2022test}. Such distribution shifts are common in real-world applications, leading to a degradation in model reliability. To mitigate this, Test-Time Adaptation (TTA) has emerged as a crucial strategy, enabling models to dynamically adapt to unlabeled test samples on the fly, thereby enhancing robustness while preserving the benefits of zero-shot learning \cite{wang2020tent, sun2020test, liu2021ttt++}. Episodic TTA, which adapts the model for each individual test sample, is particularly relevant for VLMs operating in diverse and unpredictable environments \cite{shu2022test, farina2024frustratingly}.

Current TTA approaches for VLMs often focus on optimizing learnable components. Test-Time Prompt Tuning (TPT) \cite{shu2022test} and its variants \cite{yoon2024c, feng2023diverse} adapt textual prompts by minimizing objectives like marginal entropy over augmented views of a test sample. Although effective, these methods typically require backpropagation through the large text encoders of VLMs, leading to considerable computational overhead and increased memory usage during inference \cite{sui2024just}. Other strategies involve parameter-efficient fine-tuning techniques such as Low-Rank Adaptation (LoRA) \cite{hu2022lora, imam2025test} applied to parts of the VLM. However, these approaches often necessitate access to and modification of the model's internal architecture, deviating from a truly black-box paradigm and potentially limiting their applicability to proprietary models or those with fixed structures. The challenge remains to develop TTA methods that are both highly efficient and minimally invasive while effectively addressing domain shifts.

To address these limitations, we introduce Spectrum-Aware Test-Time Steering (STS), a novel TTA framework for VLMs that operates by efficiently adapting text representations within a low-dimensional subspace defined by their Singular Value Decomposition (SVD). Instead of learning prompt vectors or modifying encoder weights, STS pre-computes a semantic basis from the SVD of the initial class text embeddings. At test time, for each incoming sample, our method learns a small set of coefficients that define a shift vector within this principal SVD subspace. This shift is then applied to the initial text prototypes, effectively steering them in the joint embedding space to better align with the current visual input. This approach directly manipulates representations in the latent space in a highly targeted and parameter-efficient manner.

The core strength of STS lies in its strategy of adapting representations within a structured, low-dimensional subspace, a design motivated by the observation that embeddings derived from pre-trained deep neural networks are typically characterized by a low intrinsic dimension \cite{aghajanyan2020intrinsic}. This implies that their essential information resides within a lower-dimensional manifold, which can be effectively identified through spectral decomposition methods such as SVD. By operating on the principal singular vectors derived from the initial text embeddings, our method explicitly leverages the inherent semantic geometry of the VLM's text feature space. These singular vectors capture the most salient axes of variation among class concepts, providing a robust and semantically grounded basis for adaptation. Performing test-time adaptation within this constrained subspace inherently regularizes the learning process, fostering enhanced stability against noisy augmentations or idiosyncrasies of individual test samples. This targeted manipulation not only preserves the rich knowledge encoded in the frozen VLM but also ensures that adaptations are focused along directions of maximal semantic relevance, rather than allowing unconstrained shifts in the high-dimensional embedding space.

This work has four primary contributions:
\begin{itemize}
    \item We propose STS, a novel TTA method that, to the best of our knowledge, is the first to leverage the SVD-defined latent subspace of text embeddings for efficient and effective adaptation of VLMs to unlabeled test data.
    \item Our method exhibits significant computational advantages: it avoids backpropagation through the VLM encoders and adapts only a minimal number of parameters (the SVD subspace coefficients), leading to substantially lower latency and memory footprint compared to conventional prompt tuning techniques.
    \item STS operates as a black-box adaptation mechanism, treating the VLM encoders as fixed feature extractors without requiring knowledge of or modifications to their internal architectures. This makes our approach broadly applicable and non-invasive.
    \item Extensive experiments on several benchmark datasets for natural distribution shifts and cross-dataset generalization demonstrate that STS achieves state-of-the-art performance, enhancing the zero-shot capabilities of VLMs efficiently and effectively.
\end{itemize}
\section{Related work}
\label{sec:relatedwork}

\paragraph{Vision-Language Models.} Pre-trained on extensive image-text datasets via self-supervised learning, vision-language models (VLMs) like CLIP \cite{radford2021learning} and ALIGN \cite{jia2021scaling} exhibit impressive generalization abilities. For instance, CLIP’s exceptional zero-shot performance largely stems from the scale and variety of its training data. However, effectively adapting these models to specific downstream tasks, especially in data-scarce scenarios, continues to pose significant challenges. Efforts to improve the transferability of vision-language models have led to the use of prompt tuning techniques, including CoOp \cite{zhou2022learning}, CoCoOp \cite{zhou2022conditional}, and MaPLe \cite{khattak2023maple}, and adapter-based methods, such as Tip-Adapter \cite{zhang2022tip} and CLIP-Adapter \cite{gao2024clip}. However, a prevalent assumption in these techniques is the accessibility of labeled data from the target domain, a condition that is frequently incompatible with the requirements for rapid deployment in real-world applications. Therefore, we focus on test-time adaptation, defined as the challenge of a model adapting to a target domain using only the unlabeled test instances, with no access to any training data or ground-truth labels from that specific domain.

\paragraph{Test-Time Adaptation.} Test-Time Adaptation (TTA) aims to improve model robustness and generalization by adapting a pre-trained model to unlabeled test data encountered during inference \cite{, sun2020test, liu2021ttt++, wang2020tent, gao2022visual, mirza2023actmad, chroni2024improving, shu2022test}. In the context of VLMs, several TTA strategies have emerged.

A dominant paradigm involves tuning the learnable prompt vectors at test time. Test-Time Prompt Tuning (TPT) \cite{shu2022test} pioneered this by optimizing textual prompts for each test sample to minimize the entropy of predictions over augmented views. Subsequent works have built upon TPT, such as DiffTPT \cite{feng2023diverse}, which employs diffusion models for more diverse augmentations, and C-TPT \cite{yoon2024c}, which focuses on improving model calibration during TTA. Although these methods adapt VLMs without labeled data, they generally incur significant computational costs and memory overhead due to the need for backpropagation through the large VLM encoders to update the prompt parameters.

To address the efficiency concerns of prompt tuning, training-free approaches have been proposed. Most of these methods operate in an on-line streaming scenario, using memory banks to retain information from previously seen test inputs \cite{zhang2024dual, karmanov2024efficient}. However, such methods suffer from critical drawbacks that limit their applicability to real-life scenarios. First, maintaining memory banks significantly increases memory consumption, which becomes prohibitive for resource-constrained devices or large-scale deployments. Second, their efficacy hinges on the assumption of a well-distributed stream of data, an unrealistic expectation in practice, as imbalanced or nonstationary test distributions may prevent the memory bank from accumulating sufficient or representative samples within a reasonable time-frame. This dependency renders them unreliable for time-sensitive applications or scenarios with sparse or bursty data streams. Furthermore, memory-based methods risk performance degradation when test samples arrive in biased sequences, as stored information may reflect transient patterns rather than meaningful statistical trends. Our STS method, while involving a lightweight optimization step, shares the goal of minimizing encoder backpropagation and architectural changes.

Other approaches explore parameter-efficient fine-tuning (PEFT) techniques at test time. For example, TTL \cite{imam2025test} adapts LoRA \cite{hu2022lora} parameters within the VLM's attention layers. Although more efficient than full fine-tuning, such methods still typically require modification of the underlying model architecture, differing from black-box approaches where the VLM encoders are treated as fixed.

Directly adapting or modulating representations in the latent embedding space offers an alternative to prompt tuning or architectural modifications. Test-Time Prototype Shifting (TPS) \cite{sui2024just} proposes learning shift vectors for pre-computed class prototypes directly in the embedding space, thereby avoiding backpropagation through encoders and achieving significant efficiency gains. This is conceptually related to our work. However, TPS learns unconstrained shift vectors for each class prototype. Our proposed STS method advances the idea of latent space adaptation by introducing a spectrum-aware mechanism. Instead of learning arbitrary shifts, STS learns compact coefficients that define shifts along principal semantic axes derived from the SVD of text embeddings. This constrains the adaptation to a low-dimensional, semantically meaningful subspace, aiming for both efficiency and effectiveness.

\begin{figure*}[!t]
 \centering
\includegraphics[width=1.0\linewidth]{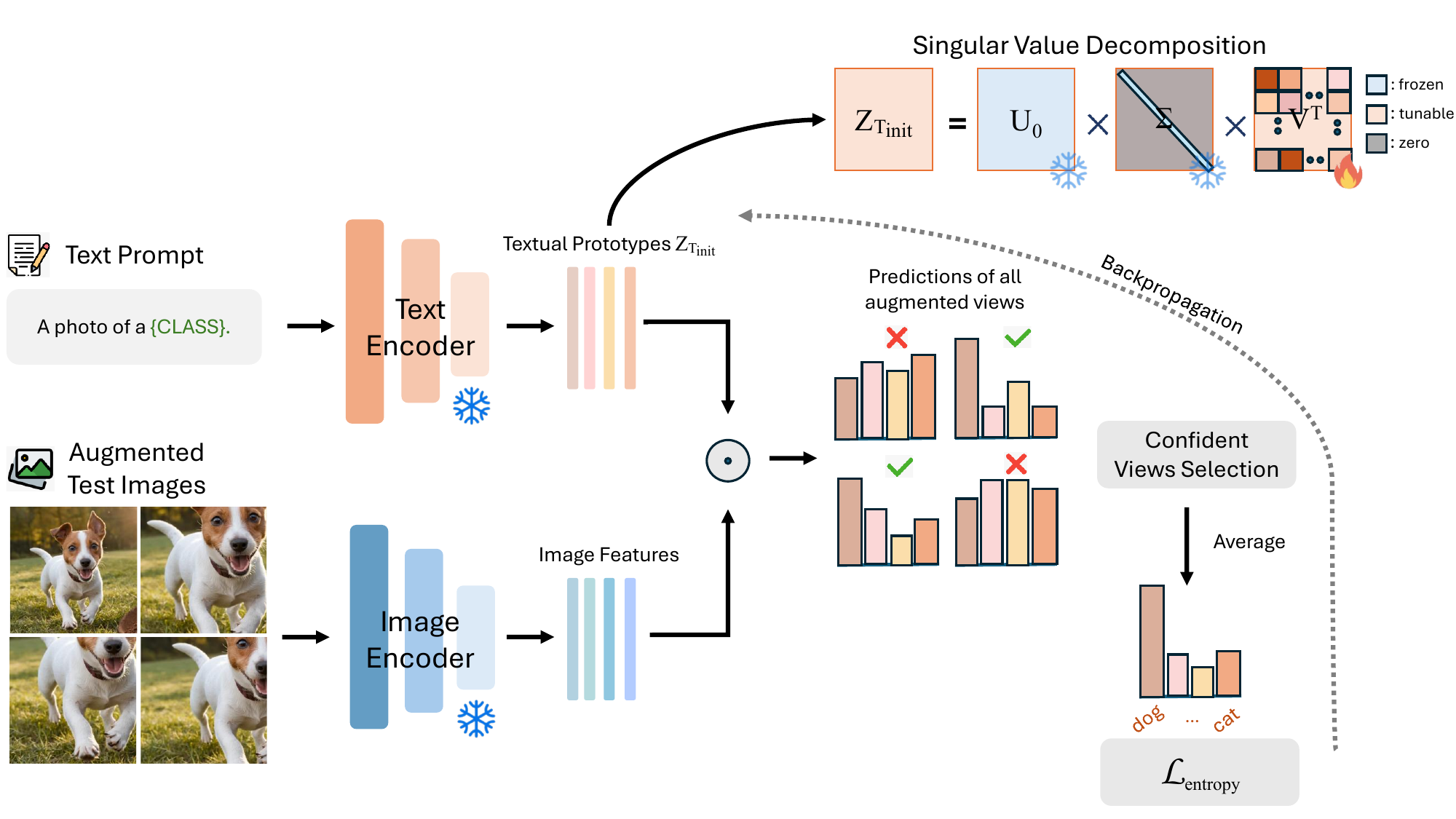}
\caption{Overview of our proposed STS framework. Given text and image inputs, encoders $\mathcal{E}_t(\cdot)$ and $\mathcal{E}_v(\cdot)$ extract text embeddings/prototypes, and visual embeddings. A probability distribution $\mathbb{P}_{CLIP}(y=y_c|X_{\mathtt{test}})$ is computed based on these embeddings. Then we perform a refinement step of test-time adaptation, where we tune the learnable low-dimensional coefficients to generate a small steering to the text prototypes to close the gap between the source and target distributions. Marginal entropy of the CLIP similarities of the shifted embeddings and the class prototypes is minimized.}
\label{fig:pipeline}  
\vspace{-10pt}
\end{figure*}

\section{Method}
\label{sec:method}

In this section, we first provide preliminary details on Vision-Language Models (VLMs) and the test-time adaptation setting.
Subsequently, we introduce our proposed \textbf{S}pectrum-Aware \textbf{T}est-Time \textbf{S}teering (STS) framework. We detail its components for identifying a principal spectral subspace from text embeddings, the mechanism for latent steering via a shared, learnable coefficient vector operating within this subspace, and the test-time optimization objective.

\subsection{Preliminaries}

\paragraph{Vision-Language Models (VLMs).}
Pre-trained VLMs, such as CLIP~\cite{radford2021learning}, comprise a visual encoder $\mathcal{E}_v(\cdot)$ and a textual encoder $\mathcal{E}_t(\cdot)$, which map images $x$ and text descriptions $t$ into a shared $D$-dimensional embedding space.
For a $C$-class zero-shot image classification task, a set of initial text prototypes, $Z_{T_{\text{init}}} = \{ (z_{T_{\text{init}}})_c \}_{c=1}^C \in \mathbb{R}^{C \times D}$, is derived by encoding class names, typically using prompt templates (e.g., "a photo of a \texttt{\{class name\}}").
Given an input image $x$, its visual embedding is $z_v = \mathcal{E}_v(x) \in \mathbb{R}^{D}$.
The predicted probability for class $c$ is:
\begin{equation}
p(y=c | x, Z_{T_{\text{init}}}) = \frac{\exp(\text{sim}(z_v, (z_{T_{\text{init}}})_c) / \tau)}{\sum_{i=1}^{C} \exp(\text{sim}(z_v, (z_{T_{\text{init}}})_i) / \tau)},
\label{eq:clip_prediction}
\end{equation}
where $\text{sim}(\cdot, \cdot)$ denotes cosine similarity, and $\tau$ is CLIP's learned temperature.

\paragraph{Test-Time Adaptation (TTA).}
Episodic TTA seeks to adapt a model $f$ for each unlabeled test sample $x_{\text{test}}$ by optimizing a small set of parameters $\theta_{\text{TTA}}$ using an objective $\mathcal{L}_{\text{TTA}}$ derived from $x_{\text{test}}$ (typically via augmentations). The adapted model $f(\cdot; \theta_{\text{TTA}}^*)$ is then used for prediction, and $\theta_{\text{TTA}}$ is reset for subsequent samples.

\subsection{Spectrum-Aware Test-Time Steering (STS)}
STS adapts VLMs at test time by \textit{learning to steer} the initial text prototypes. This steering is enacted by a single, shared vector of learnable coefficients that operates within a low-dimensional semantic subspace. This subspace is critically derived from the spectral properties (Singular Value Decomposition) of the initial text embeddings, ensuring adaptations are both efficient and aligned with principal semantic variations.

\subsubsection{Spectral Subspace Identification from Text Prototypes}
\label{sec:spectral_subspace_identification}
The core of STS lies in identifying a robust, low-dimensional subspace from the initial text prototypes $Z_{T_{\text{init}}} \in \mathbb{R}^{C \times D}$ to guide the adaptation. High-dimensional embedding spaces, while expressive, can be susceptible to noise and may contain redundant information for specific adaptation tasks. Pre-trained features from deep neural networks often exhibit a low intrinsic dimensionality~\cite{aghajanyan2020intrinsic}, implying that their essential information can be effectively captured within a lower-dimensional manifold. By projecting the adaptation process onto such a manifold, defined by the principal spectral components, STS aims to achieve more stable, generalizable, and semantically meaningful adaptations.

To this end, we perform a Singular Value Decomposition (SVD) on the initial text prototypes $Z_{T_{\text{init}}}$. Using the reduced SVD:
\begin{equation}
Z_{T_{\text{init}}} = U_T S_T V_T^\top,
\end{equation}
where $U_T \in \mathbb{R}^{C \times k'}$ contains the left singular vectors, $S_T \in \mathbb{R}^{k' \times k'}$ is a diagonal matrix of $k' = \min(C,D)$ singular values $s_1 \ge s_2 \ge \dots \ge s_{k'} \ge 0$, and $V_T^\top \in \mathbb{R}^{k' \times D}$ has rows corresponding to the right singular vectors. The columns of $V_T \in \mathbb{R}^{D \times k'}$ (i.e., the right singular vectors of $Z_{T_{\text{init}}}$) represent an orthonormal basis for the subspace capturing the principal directions of variance in the text prototype data. These directions correspond to the most significant semantic axes that differentiate the class concepts as represented by $Z_{T_{\text{init}}}$.

\begin{figure*}[!t]
 \centering
\includegraphics[width=1.0\linewidth]{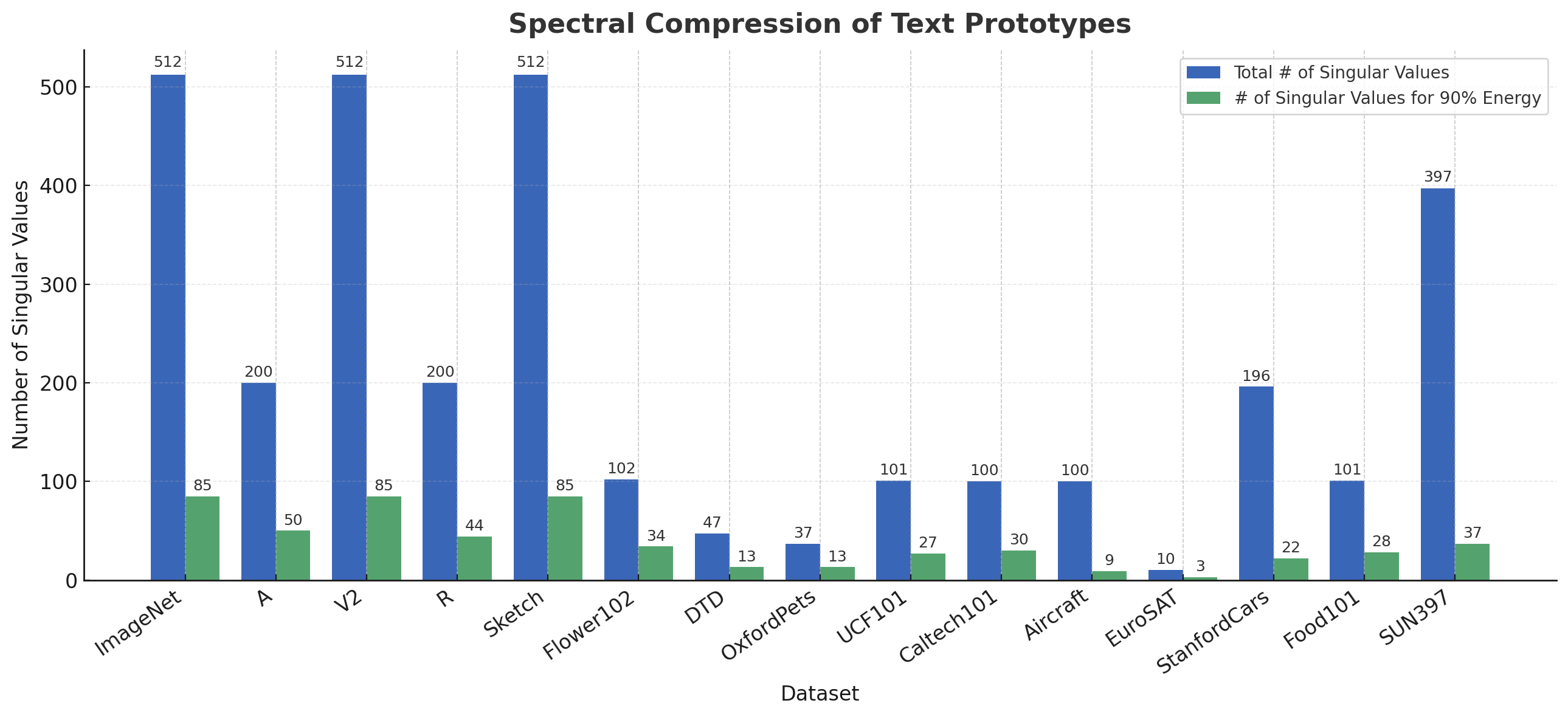}
\caption{Most spectral energy in CLIP text prototypes is captured by a small subset of singular values, highlighting strong low-rank structure across datasets.}
\label{fig:ranks}  
\end{figure*}

We select the top $k_t$ right singular vectors from $V_T$ (associated with the $k_t$ largest singular values) to form our textual adaptation basis $B_T = [v_1, v_2, \dots, v_{k_t}] \in \mathbb{R}^{D \times k_t}$. The choice of $k_t \ll D$ is pivotal for efficiency and robustness. A small $k_t$ focuses adaptation on the most dominant semantic variations, potentially filtering out noise associated with higher-order components and leveraging the aforementioned low intrinsic rank of deep features \cite{aghajanyan2020intrinsic}. Empirically, as we see in Figure~\ref{fig:ranks}, a small $k_t$ often captures the vast majority (e.g., >90\%) of the total variance (sum of squared singular values, or "energy") of $Z_{T_{\text{init}}}$. Thus, for automatic and principled selection of $k_t$, we employ the optimal hard thresholding strategy proposed by Gavish and Donoho \cite{gavish2017optimal}. This method determines an optimal singular value threshold $\omega^*$ based on the aspect ratio of the matrix $Z_{T_{\text{init}}}$ (i.e., $C/D$) and the median of its singular values. The rank $k_t$ is then the count of singular values $s_i$ such that $s_i > \omega^*$. This pre-computed basis $B_T$ defines the $k_t$-dimensional spectral subspace for our test-time steering.

\subsubsection{Latent Steering via Subspace Coefficients}
For each incoming test image $x_{\text{test}}$, STS learns a \textit{single, shared} vector of $k_t$ learnable coefficients $\mathbf{\gamma} \in \mathbb{R}^{k_t}$. These coefficients determine the magnitude and direction of the text prototype steering along each of the $k_t$ basis vectors in $B_T$.
The steering vector (shift) $\Delta z_T \in \mathbb{R}^{D}$, which is applied to all class prototypes, is reconstructed from its $k_t$-dimensional representation $\mathbf{\gamma}$:
\begin{equation}
\Delta z_T = B_T \mathbf{\gamma}.
\end{equation}
The adapted text prototype $(z_{T_{\text{adapted}}})_c$ for each class $c$ is then obtained by:
\begin{equation}
(z_{T_{\text{adapted}}})_c = \text{normalize} \left( (z_{T_{\text{init}}})_c + \Delta z_T \right).
\label{eq:adapted_text_prototype}
\end{equation}
The only parameters learned at test time are the $k_t$ coefficients in $\mathbf{\gamma}$.

\subsubsection{Test-Time Optimization Objective}
The shared steering coefficients $\mathbf{\gamma}$ are optimized for each $x_{\text{test}}$ using an unsupervised objective based on prediction consistency over $N$ augmented views of $x_{\text{test}}$, denoted $\{x^{(j)}\}_{j=1}^N$. Visual embeddings $Z_V = \{ z_v^{(j)} = \mathcal{E}_v(x^{(j)}) \}_{j=1}^N$ are extracted using the frozen $\mathcal{E}_v$.

\paragraph{Confidence Filtering.}
Following prior TTA works~\cite{shu2022test, farina2024frustratingly}, views are filtered based on prediction confidence using the initial (unadapted) text prototypes $Z_{T_{\text{init}}}$. Logits $L_{\text{init}}^{(j)}$ and probabilities $P_{\text{init}}^{(j)}$ are computed per view. Views with prediction entropy $H(P_{\text{init}}^{(j)})$ falling within the top-$\rho$ percentile of confidence (lowest entropy) are retained, forming $Z_{V_{\text{filt}}} = \{ z_v^{(j')} \}$ of size $N_{\text{filt}}$.

\paragraph{Marginal Entropy Minimization.}
For $z_v^{(j')} \in Z_{V_{\text{filt}}}$, logits with current adapted text prototypes $Z_{T_{\text{adapted}}}$ (from Eq.~\ref{eq:adapted_text_prototype} using current $\mathbf{\gamma}$) are:
$L_{\text{adapted}}^{(j')}(c) = \text{sim}(z_v^{(j')}, (Z_{T_{\text{adapted}}})_c) / \tau$.
The marginal probability distribution $\bar{P}_{\text{adapted}}$ is:
\begin{equation}
\bar{P}_{\text{adapted}}(c) = \frac{1}{N_{\text{filt}}} \sum_{j'=1}^{N_{\text{filt}}} \text{softmax}_c(L_{\text{adapted}}^{(j')}(c)).
\end{equation}
The primary objective is to minimize the Shannon entropy of $\bar{P}_{\text{adapted}}$:
\begin{equation}
\mathcal{L}_{\text{ent}} = H(\bar{P}_{\text{adapted}}) = - \sum_{c=1}^{C} \bar{P}_{\text{adapted}}(c) \log \bar{P}_{\text{adapted}}(c).
\end{equation}
An L2 regularization term is added for $\Delta z_T$: $\mathcal{L}_{\text{reg}} = \lambda_R ||\Delta z_T||_2$.
The total loss is:
\begin{equation}
\mathcal{L}_{\text{STS}} = \mathcal{L}_{\text{ent}} + \mathcal{L}_{\text{reg}}.
\end{equation}
The coefficients $\mathbf{\gamma}$ are initialized to zeros and optimized to minimize $\mathcal{L}_{\text{STS}}$.

\subsubsection{Inference with Adapted Prototypes}
After optimization yielding $\mathbf{\gamma}^*$, the final adapted text prototypes $Z_{T_{\text{final}}}$ are computed. The final class prediction $\hat{y}$ for $x_{\text{test}}$ is:
\begin{equation}
\hat{y} = \text{argmax}_c \, \left( \frac{1}{N_{\text{filt}}} \sum_{j'=1}^{N_{\text{filt}}} \text{softmax}_c \left( \text{sim}(z_v^{(j')}, (Z_{T_{\text{final}}})_c) / \tau \right) \right).
\end{equation}

\section{Experiments and Results}
\label{sec:experiments}
We conduct experiments on a diverse range of benchmark datasets to assess the performance and robustness of our method, specifically testing its out-of-domain generalization across different domains.

\subsection{Experimental Setup}
\paragraph{Datasets.} We conduct a comprehensive evaluation of our method across a diverse set of benchmark datasets, with a particular focus on out-of-domain generalization. To assess the model’s ability to handle distribution shifts, we utilize several ImageNet variants, including ImageNet-A \cite{hendrycks2021natural}, ImageNet-V2 \cite{recht2019imagenet}, ImageNet-R \cite{hendrycks2021many}, and ImageNet-Sketch (also referred to as ImageNet-K) \cite{wang2019learning}. These datasets serve as established Out-of-Distribution (OOD) benchmarks for ImageNet, enabling a rigorous assessment of the model’s robustness under varying conditions and data distributions.

For Fine-grained Classification (also referred to as "Cross-Datasets Generalization in previous works), in line with \cite{shu2022test}, we include Flowers102 \cite{nilsback2008automated}, DTD \cite{cimpoi2014describing}, Pets \cite{parkhi2012cats}, UCF \cite{soomro2012ucf101}, and Caltech101 \cite{fei2004learning}. These datasets facilitate the evaluation of the model’s capacity to distinguish fine-grained variations among visually similar classes. Furthermore, to assess the model’s adaptability across diverse domains, we incorporate Aircraft \cite{maji2013fine}, EuroSAT \cite{helber2019eurosat}, Cars \cite{krause20133d}, Food \cite{bossard2014food}, and SUN397 \cite{xiao2010sun}, encompassing a broad spectrum of data modalities, including aerial and satellite imagery, object-centric datasets, and scene-centric environments. For all datasets, we utilize the test splits defined by Zhou et al. \cite{zhou2022learning}, adhering to the common evaluation protocol.

\paragraph{Implementation Details.} Following TPT \cite{shu2022test}, we generate 63 augmented versions of a test image using \textit{only} random resized crops and horizontal flips \textit{for all datasets}, unlike prior methods that use task-specific augmentations such as AugMix, resulting in a batch of 64 images including the original input. To identify high-confidence samples, we select the 10\% of batch samples with the lowest entropy and compute the marginal entropy based on their predicted probability distributions. The learnable vector is initialized to zero and optimized for a single step using the AdamW \cite{loshchilov2017decoupled} optimizer with a learning rate of 5e-3 across all datasets. In our method, each class prototype is initialized using the hand-crafted prompt, “a photo of a \texttt{\{CLASS\}}.” All experiments are conducted on a single NVIDIA RTX8000 GPU with 45GB of memory.  The presented results are an average taken over three distinct random seeds. Top-1 accuracy is reported in all tables, unless explicitly indicated otherwise.

\paragraph{Baselines.} We evaluate our method against zero-shot and test-time adaptation (TTA) baselines that utilize CLIP ViT-B/16 as the backbone. The TTA methods considered include TPT \cite{shu2022test}, which performs text-prompt tuning; DiffTPT \cite{feng2023diverse}, a variant of TPT that employs diffusion models to augment visual training data; TPS \cite{sui2024just}, which optimizes a shift vector for each class prototype; and C-TPT \cite{yoon2024c}, an extension of TPT that improves model calibration by selecting prompts based on the dispersion of textual embeddings. To ensure a fair comparison, we reproduce TPT, TPS, and C-TPT on our system using a single update step and the same backbone architecture. For DiffTPT, we report results from \cite{feng2023diverse}. It is important to note that the DiffTPT study evaluates performance on a subset of each dataset that contain only 1,000 test samples, which may introduce potential variability in the reported results.

\paragraph{Textual Prompts.} When \textit{Ensemble} is specified, we do not use dataset-specific templates. In contrast, we use the set of 7 generic templates highlighted in the official CLIP repository \cite{clip2021github} across all datasets.

\begin{table*}[t]
\caption{Comparison of top-1 accuracy (\%) across ImageNet and its OOD variants. The best results in each section are highlighted in \textbf{bold}. \underline{Underline} indicates second-best.}
\label{tab:natural_and_results} 
\centering
\begin{tabular}{l c c c c c c c }
\toprule
\textbf{Method} & \textbf{ImageNet} & \textbf{A} & \textbf{V2} & \textbf{R} & \textbf{Sketch} & \textbf{Average} & \textbf{OOD Average} \\
\midrule
\rowcolor{lightergray}
\multicolumn{8}{c}{\textbf{\texttt{CLIP-ViT-B/16}}} \\
\midrule
Zero-Shot \cite{radford2021learning} & 66.73 & 47.87 & 60.86 & 73.98 & 46.09 & 59.11 & 57.2 \\
\midrule
Ensemble \cite{clip2021github}  & 68.34 & 49.89 & 61.88 & 77.65 & 48.24 & 61.20 & 59.42 \\
CoOp \cite{zhou2022learning} & 71.51 & 49.71 & 64.20 & 75.21 & 47.99 & 61.72 & 59.28 \\
\midrule
TPT  \cite{shu2022test}     & \underline{68.97} & 54.39 & 63.37 & 77.07    & 48.01 & 62.36 & 60.71 \\
DiffTPT  \cite{feng2023diverse}     & 70.30 & 55.68 & \textbf{65.10} & 75.00     & 46.80 & 62.28 & 60.52 \\
C-TPT  \cite{yoon2024c}     & 68.53 & 51.14 & 62.13 & 75.66     & 47.37 & 60.97 & 59.08 \\
TPS   \cite{sui2024just}    & 67.96 & 57.46 & 62.95 & 74.90     & 46.03 & 61.86 & 60.34 \\
\rowcolor{lightcyan} STS (Ours) & 68.85 & \underline{61.23} & 64.15 & \underline{77.13} & \underline{48.06} & \underline{63.88} & \underline{62.64} \\
\rowcolor{lightcyan} STS$_{\text{Ensemble}}$ & \textbf{70.81} & \textbf{64.29} & \underline{64.82} & \textbf{80.53} & \textbf{50.19} & \textbf{66.13} & \textbf{64.96} \\
\midrule
\rowcolor{lightergray}
\multicolumn{8}{c}{\texttt{\textbf{MaPLe}}} \\
\midrule
Zero-Shot \cite{khattak2023maple}  & - & 50.90 & 64.07 & 76.98     & \underline{49.15} &  - & 60.28 \\
\midrule
TPT   \cite{shu2022test}    & - & \underline{58.08} & \underline{64.87} &  \underline{78.12}   & 48.16 &  -
& \underline{62.31} \\
\rowcolor{lightcyan} STS (Ours) & - & \textbf{64.83} & \textbf{66.49} & \textbf{79.43} & \textbf{50.62} & - & \textbf{65.34} \\
\bottomrule
\end{tabular}
\end{table*}

\begin{table*}[t]
\caption{Performance comparisons on fine-grained classification. The best results in each section are highlighted in \textbf{bold}. \underline{Underline} indicates second-best.}

\centering
\label{tab:natural}
\resizebox{\textwidth}{!}{  
\begin{tabular}{l c c c c c c c c c c c}
\toprule
\textbf{Method} & \textbf{Flowers102} & \textbf{DTD} & \textbf{OxfordPets} & \textbf{UCF101} & \textbf{Caltech101} & \textbf{Aircraft} & \textbf{EuroSAT} & \textbf{StanfordCars} & \textbf{Food101} & \textbf{SUN397} & \textbf{Average} \\
\midrule
\rowcolor{lightergray}
\multicolumn{12}{c}{\texttt{\textbf{CLIP-ViT-B/16}}} \\ 
\midrule
Zero-Shot \cite{radford2021learning} & 67.44 & 44.27 & 88.25 & 65.13 & 93.35 & 23.67 & 42.01 & 65.48 & 83.65 & 62.59 & 63.58 \\
\midrule
Ensemble \cite{clip2021github} & 66.99 & 45.04 & 86.92 & 65.16 & 93.55 & 23.22 & 50.42 & 66.11 & 82.86 & 65.63 & 64.59 \\
CoOp \cite{zhou2022learning} & 68.71 & 41.92 & 89.14 & 66.55 & 93.70 & 18.47 & 46.39 & 64.51 & 85.30 & 64.15 & 63.88 \\
\midrule
TPT    \cite{shu2022test}   & \underline{68.98} & \textbf{47.16} & 87.07 & \textbf{67.89} & \underline{94.19} & 22.85 & \underline{43.01} & 66.55 & 84.67 & \underline{65.47} & \underline{64.78} \\
C-TPT  \cite{yoon2024c}   & \textbf{69.88} & 45.54 & \textbf{87.96} & 65.19 & 93.39 & 24.13 & 38.43 & 65.26 & 82.60 & 63.38 & 63.58 \\
TTL   \cite{imam2025test}    & 67.32 & 45.92 & 86.78 & \underline{67.80} & 93.23 & 24.00 & 36.52 & 65.95 & 84.40 & 64.02 & 63.59 \\
TPS  \cite{sui2024just}    & 66.14 & 45.49 & 86.56 & 66.53 & 93.60 & 24.01 & 37.85 & 66.93 & 82.96 & 64.85 & 63.49 \\
\rowcolor{lightcyan} STS (Ours)  & 66.10 & 46.02 & 86.69 & 66.52 & 93.72 & \textbf{24.57} & 38.26 & \underline{67.17} & \underline{84.72} & 64.79 & 63.86 \\
\rowcolor{lightcyan} STS$_{\text{Ensemble}}$ & 67.16 & \underline{46.87} & \underline{87.11} & 67.14 & \textbf{94.20} & \underline{24.21} & \textbf{43.80} & \textbf{68.16} & \textbf{85.15} & \textbf{66.79} & \textbf{65.06} \\
 \midrule
 \rowcolor{lightergray}
\multicolumn{12}{c}{\texttt{\textbf{MaPLe}}} \\ 
\midrule
Zero-Shot \cite{khattak2023maple} & \underline{72.23} & \underline{46.49} & \underline{90.49} & 68.69 & 93.53 & \underline{24.74} & \textbf{48.06} & 65.57 & 86.20 & 67.01 & 66.30 \\
\midrule
TPT  \cite{shu2022test}     & \textbf{72.37} & 45.87 & \textbf{90.72} & \textbf{69.19} & \underline{93.59} & 24.70 & \underline{47.80} & \underline{66.50} & \textbf{86.64} & \textbf{67.54} & \textbf{66.49} \\
\rowcolor{lightcyan} STS (Ours) & 70.70 & \textbf{47.60} & 90.00 & \underline{68.94} & \textbf{94.02} & \textbf{25.44} & 40.83 & \textbf{68.32} & \underline{86.56} & \underline{67.26} & \underline{65.97} \\
\bottomrule
\end{tabular}
}
\end{table*}

\subsection{STS Results}
\paragraph{Natural Distribution Shifts.} 
Table \ref{tab:natural_and_results} presents the top-1 accuracy of our method, comparing it against zero-shot and test-time adaptation (TTA) baselines using CLIP on ImageNet and its out-of-distribution (OOD) variants. The results demonstrate that steering the text prototypes with our learnable vector, leads to a substantial performance improvement. Specifically, our approach achieves an average OOD performance gain of 7.76\% over the zero-shot CLIP baseline and a 4.23\% improvement over standard TPT across OOD datasets. 

Furthermore, Table \ref{tab:natural_and_results} shows that simply learning a shift vector for per-class prototypes in TPS results in a slight performance drop ($<0.6$ points) compared to TPT, highlighting the limitations of TPS in effectively aligning text prototypes with out-of-distribution visual embeddings. Additionally, our method is significantly more efficient, as STS runs 8 times faster than TPT (see Table \ref{tab:comparison}) while still achieving superior performance. These substantial speed-ups make our approach highly practical for real-world applications.

\begin{table*}[h]
\caption{Efficiency comparison on ImageNet. We report the testing time per sample, the memory usage, the accuracy, and the performance gains compared to zero-shot CLIP.}
\centering
\label{tab:comparison}
\begin{tabular}{l c c c c}
\toprule
\textbf{Method} & \textbf{Testing Time (s)} & \textbf{Memory (GB)} & \textbf{Accuracy} & \textbf{Gain} \\
\midrule
Zero-Shot  &  0.02 & 0.83 & 66.73  & - \\
TPT   & 0.75 & 17.6 & 68.97 & +2.24 \\
\rowcolor{lightcyan} STS$_{\text{Ensemble}}$ &  \textbf{0.09} & \textbf{1.4} & \textbf{70.81} & \textbf{+4.08} \\
\bottomrule
\end{tabular}
\end{table*}

\paragraph{Fine-grained Classification.} 
We further assess the generalization capabilities of STS across ten diverse image classification datasets, with results presented in Table~\ref{tab:natural}. This benchmark evaluates the model's ability to adapt to datasets that may differ significantly in domain and class composition from the VLM's pre-training data.

In this challenging scenario, our STS$_{\text{Ensemble}}$ variant, which leverages 7 generic CLIP templates, achieves the highest average accuracy of 65.06\% among all methods compared. This result surpasses the standard TPT~\cite{shu2022test} (average 64.78\%) and demonstrates the effectiveness of combining STS with prompt ensembling. The standard STS (using a single "a photo of a \texttt{\{CLASS\}}" prompt) achieves an average accuracy of 63.86\%, which is competitive and outperforms the Zero-Shot CLIP baseline (63.58\%). It also surpasses other TTA methods such as C-TPT~\cite{yoon2024c} (63.58\%), TTL~\cite{imam2025test} (63.59\%), and TPS~\cite{sui2024just} (63.49\%) on average. Standard STS particularly excels on datasets like Aircraft (24.57\%, best in its group), StanfordCars (67.17\%, second best), and Food101 (84.72\%, second best). The STS$_{\text{Ensemble}}$ variant shows broad strength, achieving the best results on Caltech101 (94.20\%), EuroSAT (43.80\%), StanfordCars (68.16\%), Food101 (85.15\%), and SUN397 (66.79\%). This indicates that while standard TPT is a strong baseline, STS, especially when combined with prompt ensembling, offers a more robust generalization across these diverse datasets.

Overall, STS demonstrates robust and often superior performance in fine-grained classification. The STS$_{\text{Ensemble}}$ variant, in particular, sets a new state-of-the-art average across the CLIP-ViT-B/16 backbone experiments. Even with stronger MaPLe~\cite{khattak2023maple} initializations, STS provides a more effective adaptation than TPT. These results, combined with STS's significant advantages in parameter efficiency and computational speed (detailed in Table~\ref{tab:comparison}), underscore its potential as a practical and powerful approach for real-world test-time adaptation of VLMs.

\section{Analysis and ablation}
\label{sec:ablation}
We perform ablation studies to assess the effect of key design choices on performance. For consistency, all analyses use ImageNet and ImageNet-A with the ViT-B/16 backbone. Additionally, we evaluate STS on CIFAR10-C ~\cite{hendrycks2019benchmarking} to test its robustness under challenging distribution shifts.

\subsection{Computational Analysis}
\textbf{Trainable Parameters:} Test-time tuning approaches such as TPT, DiffTPT, and C-TPT adapt textual prompts using only 2048 trainable parameters, corresponding to four tokens with d=512. However, these methods exhibit limited generalization, achieving a Top-1 accuracy of approximately 51\% to 55\% on ImageNet-A. In contrast, visual adaptation techniques, such as encoder tuning and layer normalization optimization, require a larger number of trainable parameters. Meanwhile, STS effectively balances this trade-off with just a small number of trainable parameters, achieving the highest generalization performance at 61.23\% while preserving model accessibility constraints.
Table \ref{tab:comparison} shows the testing time per sample and the performance gain on ImageNet for STS and the TTA baseline TPT on a single RTX8000 GPU. STS is 8x faster than TPT, corresponding to an order of magnitude of computational savings in time.

\subsection{Effect of STS on Different Prototypes}
We study the impact of the steering vector under different prototype constructions. Specifically, we compare STS against the zero-shot CLIP using the standard prompt \textit{"a photo of a \texttt{\{CLASS\}}"} and an ensemble of seven generic hand-crafted templates from the official CLIP repository. While the design of TPT does not support the use of text ensembles, STS integrates them seamlessly; we denote this variant as STS$_{\text{Ensemble}}$. As shown in Tables~\ref{tab:natural_and_results} and~\ref{tab:natural}, adding these generic prompts further improves STS, surpassing TPT without relying on dataset-specific templates. 

Furthermore, we evaluate our method using a MaPLe~\cite{khattak2023maple} initialization, where the MaPLe prompts are learned on ImageNet in a 16-shot setup following~\cite{abdul2023align}. In this evaluation, we also report results for TPT applied on top of MaPLe, as in~\cite{abdul2023align}, which we refer to as MaPLe+TPT. Although MaPLe+TPT performs better than previous methods that rely on hand-crafted prompts, our STS method notably outperforms MaPLe+TPT on most datasets. This demonstrates that the adaptive mechanism of STS provides consistent advantages even when initialized with optimized textual and visual prompts.

\paragraph{Natural Distribution Shifts.}
Under natural distribution shifts, our STS method consistently outperforms TPT even when the baseline initialization is MaPLe, achieving an average improvement of $+3.03\%$. We omit evaluation on ImageNet in this group, as MaPLe uses it as the source dataset for model adaptation, making the comparison unfair. For completeness, Zero-Shot MaPLe attains $70.72\%$ accuracy on ImageNet, which further improves to $72.72\%$ when adapted with STS ($+2.0\%$). STS with MaPLe (MaPLe+STS) demonstrates leading performance on several natural shift datasets, including ImageNet-A ($64.83\%$), ImageNet-V2 ($66.49\%$), ImageNet-Sketch ($50.62\%$).  

\paragraph{Fine-grained Classification.}
In fine-grained classification tasks, MaPLe+TPT shows a marginal average improvement of $+0.52\%$ over MaPLe+STS. However, this difference is primarily driven by performance on a single dataset, EuroSAT, while MaPLe+STS surpasses MaPLe+TPT on roughly half of the remaining datasets by a considerable margin. For the datasets where MaPLe+STS lags behind, the differences remain minimal. As discussed in ~\cite{farina2024frustratingly}, EuroSAT constitutes a known failure mode for many TTA methods. Its analysis suggests that the unique nature of satellite imagery demands task-specific augmentation strategies, making EuroSAT a controversial benchmark for evaluating TTA performance.

\subsection{Robustness of Linear Spectrum Steering}
We evaluate STS on CIFAR10-C at the highest corruption level (severity 5) following the TPT protocol (10\% most confident views, learning rate 0.005, and the hand-crafted prompt "a photo of a \texttt{\{CLASS\}}"). STS matches TPT within 0.05\% (Figure \ref{fig:cifar10c}), while clearly outperforming the naive per-class shifting (TPS), confirming the effectiveness of spectrum-aware subspace steering. With the seven generic CLIP templates, STS reaches 67.24\%, demonstrating strong complementarity between subspace steering and prompt ensembles. Constraining adaptation to the top-$k$ singular vectors further stabilizes learning under severe corruptions.

\subsection{Balancing Inference Efficiency and Accuracy} 
\begin{figure}[t]
  \centering
  \begin{subfigure}[t]{0.48\linewidth}
    \centering
    \includegraphics[width=\linewidth,trim=2pt 2pt 2pt 2pt,clip]{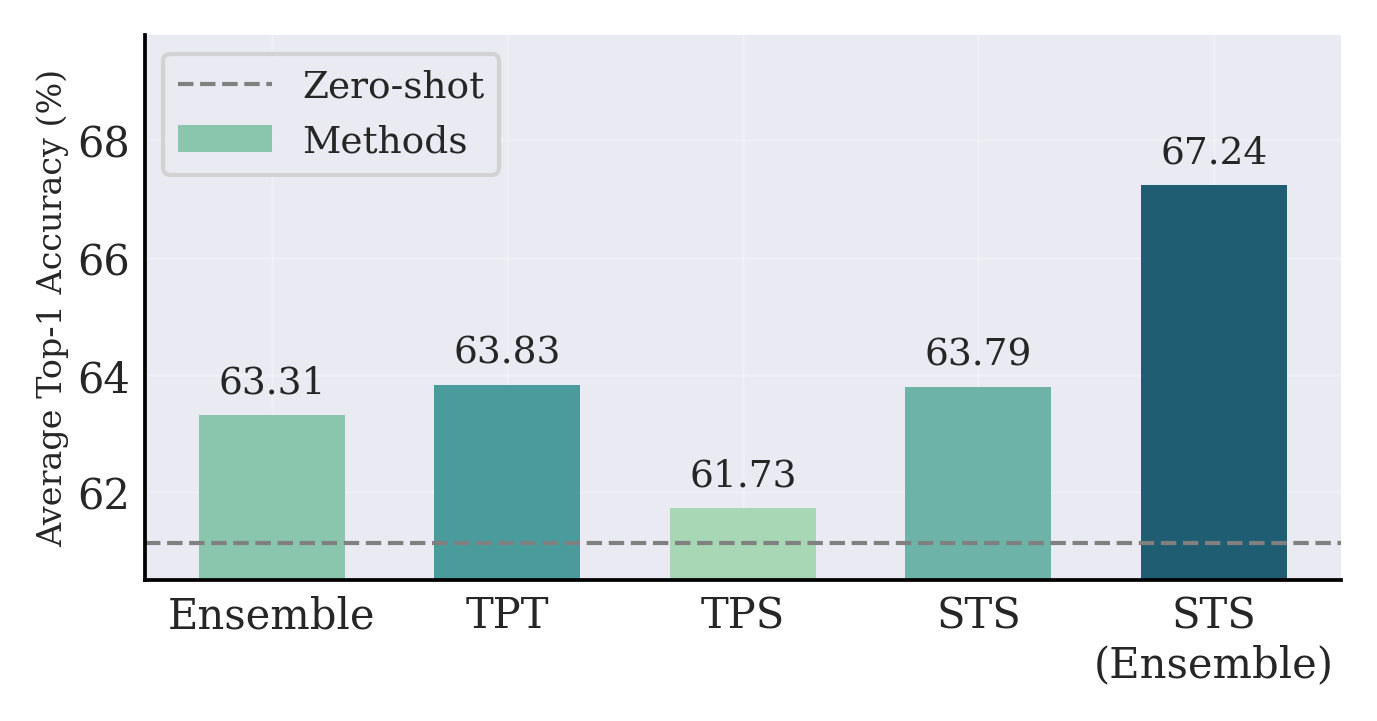}
    \caption{CIFAR10-C comparison.}
    \label{fig:cifar10c}
  \end{subfigure}
  \hfill
  \begin{subfigure}[t]{0.41\linewidth}
    \centering
    \includegraphics[width=\linewidth,trim=2pt 2pt 2pt 2pt,clip]{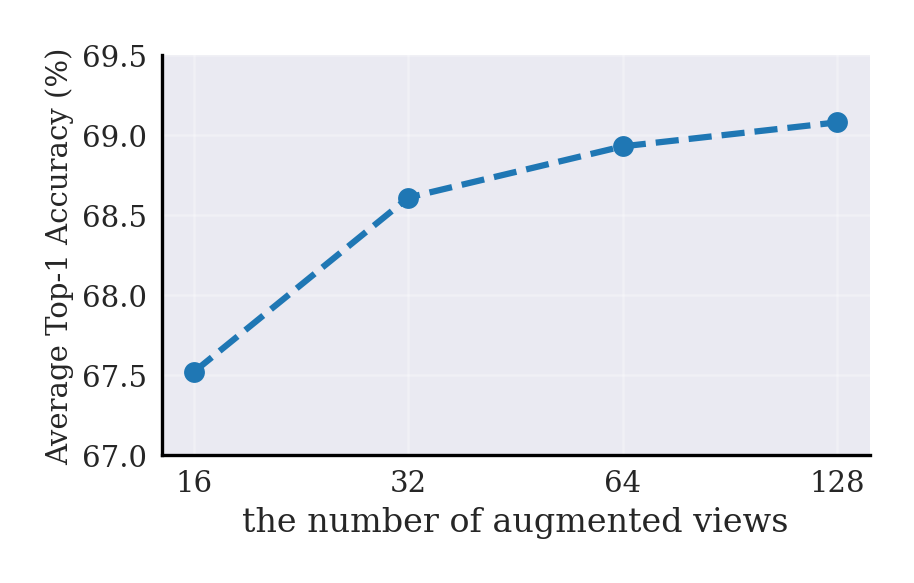}
    \caption{Accuracy vs.\ views $N$.}
    \label{fig:views}
  \end{subfigure}

  \vspace{-2mm}
  \caption{(a) Comparison on CIFAR10-C (severity 5). (b) Accuracy vs.\ number of augmented views.}
  \label{fig:fig3}
  \vspace{-3mm}
\end{figure}

We analyze the impact of the number of augmented views $N$ on STS efficiency. As shown in Figure \ref{fig:views}, accuracy increases with $N$ and saturates around $N=64$. Increasing to 128 views yields only a $\approx$0.15\% gain while nearly doubling time and memory. We thus adopt $N=64$ (as in prior TPT work) to balance performance and efficiency, while remaining substantially faster than prompt-tuning baselines.

\captionsetup[subfigure]{justification=centering,singlelinecheck=false}

\section{Limitations}
\label{sec:limitations}

While STS demonstrates significant advantages in efficiency and effectiveness for test-time adaptation, we acknowledge two limitations that warrant discussion and offer avenues for future research:

\paragraph{Linearity of subspace steering.} The adaptation mechanism involves linear shifts (steering) within the identified SVD subspace. While this subspace captures principal linear variations, highly complex or non-linear domain shifts might not be fully addressable by such linear adaptations alone, potentially requiring more sophisticated, non-linear mapping techniques within the latent space.

\paragraph{Linear complexity with respect to augmented views.} Finally, although STS is notably more lightweight than current state-of-the-art TTA strategies, its computational demand for visual processing scales linearly with the number of augmented views due to the need for independent forward passes. An intriguing future direction is to explore latent visual space augmentation to eliminate these repeated encoder computations.

By addressing these current limitations, the robustness and performance of spectrum-aware test-time adaptation strategies could be further advanced.

\section{Conclusion}
\label{sec:conclusion}

We propose Spectrum-Aware Test-Time Steering (STS), a lightweight adaptation framework for vision-language models like CLIP. STS exploits the spectral structure of text embeddings to define a compact semantic subspace, where it learns a per-sample steering vector to improve zero-shot robustness without modifying the frozen encoders. Our experiments show that STS consistently enhances performance across diverse benchmarks, offering an efficient and practical test-time adaptation method.

Notably, in addition to the text prototypes, our proposed STS method can be readily applied to the visual embeddings as well. Exploring under which conditions and settings the STS should be preferred over text prototypes or visual embeddings constitutes an interesting research direction that belongs to our future agenda.

\newpage

\section*{Acknowledgments}
We thank Konstantinos D. Polyzos for insightful discussions and for his helpful and detailed comments. We also thank the reviewers for their constructive suggestions, which have helped improve the quality of this paper. This research has been partially funded by research grants to D. Metaxas through NSF: 2310966, 2235405, 2212301, 2003874, 1951890, AFOSR 23RT0630, and NIH 2R01HL127661.

\printbibliography

\newpage
\counterwithin{table}{section}
\renewcommand{\thetable}{\thesection\arabic{table}}

\appendix

\begin{center}
\Large \textbf{Test-Time Spectrum-Aware Latent Steering for Zero-Shot Generalization in Vision-Language Models}
\end{center}
\begin{center} 
\Large Appendix
\end{center}

\vspace{2cm}

In this supplementary document, we provide additional details and experimental results to enhance
understanding and insights into our method. This supplementary document is organized as follows:
\begin{itemize}
    \item Broader Impact of our method in Section \ref{ssec:broader_impact}.
    \item We present an error bar analysis for the results in Table \ref{tab:natural_and_results} and Table \ref{tab:natural} in the main document, in Section \ref{ssec:error_bars}.
    \item We provide an analysis on the impact of varying update steps in Section \ref{update_steps}.
    \item We evaluate the effect of shared versus per-class coefficients vector in Section \ref{per_class_vs_shared}.
    \item We present additional performance comparisons on larger-scale VLMs, specifically OpenAI CLIP \cite{radford2021learning} with a ViT-L/14 backbone, in Section \ref{vitl14}.
    \item We analyze the effect of the singular vector selection for the test-time latent steering, in Section \ref{sec:svd_selection}.
    \item We provide the detailed statistics for all the utilized datasets, and the specific textual prompts that we used in Section \ref{datasets}.
    \item We list the license information for all used assets in Section \ref{licenses}.
\end{itemize}

\section{Broader Impact}
\label{ssec:broader_impact}
This research contributes to the overarching goal of developing more dependable and effective machine learning systems by enabling large foundation models like CLIP \cite{radford2021learning} to dynamically adapt to real-world operational conditions at test time. Such adaptability is critical for deploying AI robustly in diverse and unpredictable environments, thereby broadening their practical applications and fostering greater system reliability. Ultimately, we hope this work stimulates and guides future studies focused on enhancing the generalization capabilities and operational robustness of pre-trained models, ensuring they can be utilized more effectively and responsibly to address a wide array of societal challenges.

\section{Technical Appendices}

\subsection{Analysis on error bars}
\label{ssec:error_bars}
We run STS multiple times using 3 different random seeds and report the average accuracy with standard deviation in Table \ref{tab:natural_and_results_bars}. The randomness of STS mainly comes from random data augmentation. Our augmentation setup is simple
and only contains resized random crops and random horizontal flips, which can constitute a
“zoom-in” to a random portion of the image. We did not search for the best data augmentations, but rather stuck to an established setting, using the same augmentation setup for all datasets. However, the performance of STS is linked to the impact that data augmentations have on how the model perceives images, and we
believe that this is an interesting research direction to pursue. In addition, we report an error bar analysis for the results in Table \ref{tab:natural_and_results} and Table \ref{tab:natural} in the main document.

\begin{table*}[htbp]
\begingroup 
\small 
\setlength{\tabcolsep}{1.65pt} 
\caption{Robustness to natural distribution shifts. We report the accuracy with an error bar (standard deviation) obtained from three runs with different random seeds. The best results in each section are highlighted in \textbf{bold}. \underline{Underline} indicates second-best.}
\label{tab:natural_and_results_bars} 
\resizebox{\textwidth}{!}{  
\centering
\begin{tabular}{l c c c c c c c }
\toprule
\textbf{Method} & \textbf{ImageNet} & \textbf{A} & \textbf{V2} & \textbf{R} & \textbf{Sketch} & \textbf{Average} & \textbf{OOD Average} \\
\midrule
\rowcolor{lightergray}
\multicolumn{8}{c}{\texttt{\textbf{CLIP-ViT-B/16}}} \\
\midrule
TPT  \cite{shu2022test}     & \underline{68.97} {\scriptsize ($\pm .04$)}  &  54.39 {\scriptsize ($\pm .20$)}& 63.37 {\scriptsize ($\pm .06$)} &  77.07 {\scriptsize ($\pm .06$)}  &  48.01 {\scriptsize ($\pm .08$)} & 62.36 {\scriptsize ($\pm .03$)} &  60.71 {\scriptsize($\pm .04$)}\\
C-TPT   \cite{yoon2024c}    & 68.53 {\scriptsize ($\pm .02$)} & 51.14 {\scriptsize ($\pm .09$)} & 62.13 {\scriptsize ($\pm .11$)} & 75.66 {\scriptsize ($\pm .07$)} & 47.37 {\scriptsize ($\pm .08$)} & 60.97 {\scriptsize ($\pm .02$)} & 59.08 {\scriptsize ($\pm .02$)} \\
TPS  \cite{sui2024just}   & 67.96 {\scriptsize ($\pm .02$)}  & 57.46 {\scriptsize ($\pm .12$)}  & 62.95 {\scriptsize ($\pm .11$)}  & 74.90 {\scriptsize ($\pm .04$)}  & 46.03 {\scriptsize ($\pm .09$)}  & 61.86 {\scriptsize ($\pm .06$)}  & 60.34 {\scriptsize ($\pm .07$)} \\
\rowcolor{lightcyan} STS (Ours) & 68.85 {\scriptsize ($\pm .03$)} & \underline{61.23} {\scriptsize ($\pm .26$)} & 64.15 {\scriptsize ($\pm .20$)} & \underline{77.13} {\scriptsize ($\pm .06$)} & \underline{48.06} {\scriptsize ($\pm .06$)} & \underline{63.88} {\scriptsize ($\pm .08$)} & \underline{62.64} {\scriptsize ($\pm .10$)} \\
\rowcolor{lightcyan} STS$_{\text{Ensemble}}$ & \textbf{70.81} {\scriptsize ($\pm .04$)} & \textbf{64.29} {\scriptsize ($\pm .09$)} & \underline{64.82} {\scriptsize ($\pm .14$)} & \textbf{80.53} {\scriptsize ($\pm .13$)} & \textbf{50.19} {\scriptsize ($\pm .02$)} & \textbf{66.13} {\scriptsize ($\pm .01$)} & \textbf{64.96} {\scriptsize ($\pm .005$)} \\

\bottomrule
\end{tabular}
}
\endgroup 
\end{table*}

\begin{table*}[htbp]
\begingroup 
\small 
\setlength{\tabcolsep}{1.0pt} 
\caption{Performance comparisons on cross-dateset generalization from ImageNet to fine-grained classification datasets. We report the accuracy with an error bar (standard deviation) obtained from three runs with different random seeds. The best results in each section are highlighted in \textbf{bold}. \underline{Underline} indicates second-best.}
\centering
\label{tab:natural_bars}
\resizebox{\textwidth}{!}{  
\begin{tabular}{l c c c c c c c c c c c}
\toprule
\textbf{Method} & \textbf{Flowers102} & \textbf{DTD} & \textbf{OxfordPets} & \textbf{UCF101} & \textbf{Caltech101} & \textbf{Aircraft} & \textbf{EuroSAT} & \textbf{StanfordCars} & \textbf{Food101} & \textbf{SUN397} & \textbf{Average} \\
\midrule
\rowcolor{lightergray}
\multicolumn{12}{c}{\texttt{\textbf{CLIP-ViT-B/16}}} \\ 
\midrule
TPT  \cite{shu2022test}  & \underline{68.98} {\scriptsize ($\pm .13$)} & \textbf{47.16} {\scriptsize ($\pm .08$)} & 87.07 {\scriptsize ($\pm .19$)} & \textbf{67.89} {\scriptsize ($\pm .07$)} & \underline{94.19} {\scriptsize ($\pm .12$)} & 22.85 {\scriptsize ($\pm .41$)} & \underline{43.01} {\scriptsize ($\pm .06$)} & 66.55 {\scriptsize ($\pm .02$)} & 84.67 {\scriptsize ($\pm .06$)}  & \underline{65.47} {\scriptsize ($\pm .13$)}  & \underline{64.78} {\scriptsize ($\pm .05$)}\\
C-TPT  \cite{yoon2024c}   & \textbf{69.88} {\scriptsize ($\pm .19$)} & 45.54 {\scriptsize ($\pm .13$)} & \textbf{87.96} {\scriptsize ($\pm .14$)} & 65.19 {\scriptsize ($\pm .41$)} & 93.39 {\scriptsize ($\pm .16$)} & 24.13 {\scriptsize ($\pm .24$)} & 38.43 {\scriptsize ($\pm .45$)} & 65.26 {\scriptsize ($\pm .24$)} & 82.60 {\scriptsize ($\pm .16$)} & 63.38 {\scriptsize ($\pm .26$)} & 63.58 {\scriptsize ($\pm .05$)} \\
TTL   \cite{imam2025test}    & 67.32 {\scriptsize ($\pm .25$)} & 45.92 {\scriptsize ($\pm .02$)} & 86.78 {\scriptsize ($\pm .02$)} & \underline{67.80}  {\scriptsize ($\pm .06$)} & 93.23 {\scriptsize ($\pm .06$)} & 24.00 {\scriptsize ($\pm .36$)} & 36.52 {\scriptsize ($\pm .05$)} & 65.95  {\scriptsize ($\pm .24$)} & 84.40 {\scriptsize ($\pm .02$)} & 64.02 {\scriptsize ($\pm .05$)} & 63.59 {\scriptsize ($\pm .02$)} \\
TPS  \cite{sui2024just}   & 66.14 {\scriptsize ($\pm .11$)} & 45.49 {\scriptsize ($\pm .36$)} & 86.56 {\scriptsize($\pm .08$)} & 66.53 {\scriptsize($\pm .19$)} & 93.60 {\scriptsize($\pm .12$)} & 24.01 {\scriptsize($\pm .44$)} & 37.85 {\scriptsize($\pm .23$)}  & 66.93 {\scriptsize($\pm .18$)} & 82.96 {\scriptsize($\pm .09$)} & 64.85 {\scriptsize($\pm .07$)} & 63.49 {\scriptsize($\pm .07$)} \\
\rowcolor{lightcyan}  STS (Ours)  & 66.10 {\scriptsize($\pm .15$)}  & 46.02 {\scriptsize($\pm .10$)} & 86.69 {\scriptsize($\pm .18$)}  & 66.52 {\scriptsize($\pm .10$)} & 93.72 {\scriptsize($\pm .10$)} & \textbf{24.57} {\scriptsize($\pm .06$)} & 38.26 {\scriptsize($\pm .35$)} & \underline{67.17} {\scriptsize($\pm .27$)} & \underline{84.72} {\scriptsize($\pm .06$)} & 64.79 {\scriptsize($\pm .17$)} & 63.86 {\scriptsize($\pm .06$)}\\
\rowcolor{lightcyan} STS$_{\text{Ensemble}}$ & 67.16 {\scriptsize($\pm .32$)}  & \underline{46.87} {\scriptsize($\pm .12$)}  & \underline{87.11} {\scriptsize($\pm .08$)} & 67.14 {\scriptsize($\pm .06$)} & \textbf{94.20} {\scriptsize($\pm .05$)} & \underline{24.21} {\scriptsize($\pm .05$)} & \textbf{43.80} {\scriptsize($\pm .15$)} & \textbf{68.16} {\scriptsize($\pm .23$)} & \textbf{85.15} {\scriptsize($\pm .03$)} & \textbf{66.79} {\scriptsize($\pm .04$)} & \textbf{65.06} {\scriptsize($\pm .03$)}\\

\bottomrule
\end{tabular}
}
\endgroup 
\end{table*}

\newpage
\subsection{Impact of Varying Update Steps}
\label{update_steps}

\noindent 
By default, STS updates the coefficients using a single step per test instance. The optimal learning rate for this single step is determined to be 0.005 on the standard ImageNet validation set (not including any of the out-of-distribution data).  
\begin{wraptable}{r}{0.51\textwidth} 
  \centering 
  \footnotesize 
  \caption{Ablation study on different update steps in learning the steering vectors. We vary the number of update steps from 1 to 5 and report the achieved performance on ImageNet-A. Results are over 3 random seeds.} 
  \label{tab:ablation_update_steps} 
  \begin{tabular}{@{}lccccc@{}} 
    \toprule
    \# Steps & 1 & 2 & 3 & 4 & 5 \\
    \midrule
    Accuracy & 61.23 & 61.25 & 61.22 & 61.20 & 61.20 \\
    \bottomrule
  \end{tabular}
\end{wraptable}
To evaluate the impact of different numbers of update steps on overall performance, we conduct ablation experiments by varying the number of update steps from 1 to 5 and report the resulting performance on ImageNet-A. For these multi-step ablations, the value of $\lambda_R$ of the regularization loss is set to 0.01, and the initial learning rate remains 0.005. Since it is optimal for a single update step, all subsequent steps are subjected to a learning rate schedule, applying a one-time decay factor of 0.1 to the initial rate. As shown in Table~\ref{tab:ablation_update_steps}, the number of update steps does not significantly influence performance (in the range of 0.1\%). Although increasing the update steps to 2 
yields a slight performance gain of 0.02\%, it also leads to a proportional decrease in inference efficiency. Although our method is extremely efficient, given this trade-off, we adopt the single-step update as the default for balancing efficiency and performance.

\subsection{Effect of Shared vs. Per-Class Coefficients}
\label{per_class_vs_shared}
Test-time adaptation strategies vary in how they modify class representations. For instance, prompt tuning methods adjust a shared prompt that subsequently undergoes non-linear transformations via the text encoder. In the context of latent-space adaptation, class prototypes can be modulated either by individual, per-class coefficients or by a single, shared coefficients vector. A shared vector applies a uniform transformation, thus maintaining the relative geometric structure of the class prototypes after adaptation. This approach primarily targets global, dataset-level distribution shifts. Although per-class vectors could, in principle, offer finer control by providing more degrees of freedom to capture distinct class-level shifts within a domain gap, the practical benefits of such granularity warrant careful consideration. Our work investigates the efficacy of the shared coefficients vector approach, as the additional complexity introduced by per-class vectors may not consistently translate into substantial performance gains over a simpler, unified shift.

As shown in Table~\ref{tab:natural_and_results_perclass}, learning a per-class shift yields a marginal average performance increase of only 0.03\%. Similarly, Table~\ref{tab:cross_datasets_perclass} indicates that per-class coefficients provide a mere 0.01\% average improvement in cross-dataset generalization from ImageNet to fine-grained classification tasks. These minimal gains suggest that, at least for a single update step, learning per-class coefficients does not substantially enhance model performance when encountering a domain gap.

\begin{table*}[htbp]
\caption{A performance comparison of shared versus per-class steering vectors regarding robustness to natural distribution shifts. We present the average top-
1 accuracy (\%) results over 3 random seeds for a single update step. The best performance is highlighted in \textbf{bold}.}
\label{tab:natural_and_results_perclass} 
\centering
\begin{tabular}{l c c c c c c c }
\toprule
\textbf{Method} & \textbf{ImageNet} & \textbf{A} & \textbf{V2} & \textbf{R} & \textbf{Sketch} & \textbf{Average} & \textbf{OOD Average} \\
\midrule
\rowcolor{lightergray}
\multicolumn{8}{c}{\texttt{\textbf{CLIP-ViT-B/16}}} \\
\midrule
Shared    &  68.85 & 61.23 &  64.15 & 77.13 & 48.06  & 63.88  &  62.64 \\
Per-class     & \textbf{68.91}  & \textbf{61.24} & \textbf{64.20}  & \textbf{77.15}  & \textbf{48.06}  & \textbf{63.91} & \textbf{62.66} \\
\bottomrule
\end{tabular}
\end{table*}

\begin{table*}[htbp]
\caption{A performance comparison of shared versus per-class steering vectors on cross-dateset generalization from ImageNet to fine-grained classification datasets. We present the average top-
1 accuracy (\%) results over 3 random seeds for a single update step. The best performance is highlighted in \textbf{bold}.}
\centering
\label{tab:cross_datasets_perclass}
\resizebox{\textwidth}{!}{  
\begin{tabular}{l c c c c c c c c c c c}
\toprule
\textbf{Method} & \textbf{Flowers102} & \textbf{DTD} & \textbf{OxfordPets} & \textbf{UCF101} & \textbf{Caltech101} & \textbf{Aircraft} & \textbf{EuroSAT} & \textbf{StanfordCars} & \textbf{Food101} & \textbf{SUN397} & \textbf{Average} \\
\midrule
\rowcolor{lightergray}
\multicolumn{12}{c}{\texttt{\textbf{CLIP-ViT-B/16}}} \\ 
\midrule
Shared      & 66.10 & \textbf{46.02} & 86.69 & \textbf{66.52} & 93.72 & 24.57 & \textbf{38.26} & 67.17 & 84.72 & 64.79 & 63.86 \\
Per-class     & \textbf{66.15} & 45.88 & \textbf{86.71} & 66.46 & \textbf{93.78} & \textbf{24.60} & 38.23 & \textbf{67.27} & \textbf{84.77} & \textbf{64.83} & \textbf{63.87} \\
\bottomrule
\end{tabular}
}
\end{table*}

\subsection{Performance Comparison on Larger-Scale VLMs}
\label{vitl14}
Our STS method can theoretically be applied to various contrastively pre-trained vision-language
models such as CLIP ViT-B/16 and CLIP ViT-L/14. In Table \ref{tab:natural_and_results_l/14}, we use OpenAI CLIP ViT-L/14 larger-scale OpenAI CLIP model, as an example, and compare the performance of our STS method and zero-shot on robustness to natural distribution shifts. We can observe that our STS still outperforms zero-shot by a large margin on average across 5 datasets, showcasing that our method generalizes well to larger-scale VLMs.

\begin{table*}[htbp]
\caption{Performance comparison on robustness to natural distribution shifts. We present top-
1 accuracy (\%) results by employing the larger-scale ViT-L/14 variant of CLIP \cite{radford2021learning}. The reported results of STS are based on a single random seed. The best performance is highlighted in \textbf{bold}.}
\label{tab:natural_and_results_l/14} 
\centering
\begin{tabular}{l c c c c c c c }
\toprule
\textbf{Method} & \textbf{ImageNet} & \textbf{A} & \textbf{V2} & \textbf{R} & \textbf{Sketch} & \textbf{Average} & \textbf{OOD Average} \\
\midrule
\rowcolor{lightergray}
\multicolumn{8}{c}{\texttt{\textbf{CLIP-ViT-L/14}}} \\
\midrule
Zero-Shot    &  73.45 & 68.76 &  67.79 & 85.39  &  57.81  & 70.64  & 69.94  \\
\rowcolor{lightcyan} STS (Ours)      & \textbf{75.37} & \textbf{78.52} & \textbf{69.88}  & \textbf{88.07}  &  \textbf{59.85} & \textbf{74.34} & \textbf{74.08} \\
\midrule
$\Delta$    &  $+1.92$ & $+9.76$  & $+2.09$  & $+2.68$  &  $+2.04$   &  $+3.70$  &  $+4.14$  \\
\bottomrule
\end{tabular}
\end{table*}

\section{Singular Vector Selection for Test-Time Latent Steering}
\label{sec:svd_selection}

In our test-time adaptation (TTA) approach, Singular Value Decomposition (SVD) is applied to text prototypes (e.g., ``a photo of a [CLASS]'') to analyze their underlying semantic structure. The full set of singular vectors describes this structure completely. However, when adapting to a new out-of-distribution (OOD) domain at test time, not all components of this original structure may be equally relevant or contribute equally to successful adaptation. The process of selecting a subset of singular vectors aims to distill the dimensions most pertinent for characterizing the new domain's relationship to the class concepts, potentially leading to a more focused and effective adaptation. As indicated in the main document (Section 3), a substantial portion of the variance (e.g., 90\%) is often concentrated in a significantly smaller subset of singular values, highlighting the potential for effective dimensionality reduction. Discarding low-variance components here means removing directions where our specific classes are textually very similar.
 
This section details two principled methods for selecting $k_t$ singular vectors and presents their impact on the zero-shot performance of CLIP \cite{radford2021learning} when integrated with our TTA method. Our findings indicate that strategic selection of singular vectors significantly enhances performance, with the Gavish-Donoho method yielding slightly superior results.

\subsection{Energy-Based Singular Vector Selection}
\label{ssec:energy_svd}
A common heuristic for dimensionality reduction via SVD is to retain the top-$k_t$ singular values, $\sigma_1 \geq \sigma_2 \geq \dots \geq \sigma_{k'}$, such that they capture a predefined percentage of the total "energy" (sum of squared singular values). We investigate a threshold of 98\% energy, selecting $k_t$ such that:
$$ \frac{\sum_{i=1}^{k_t} \sigma_i^2}{\sum_{j=1}^{k'} \sigma_j^2} \geq 0.98 $$
This method aims to preserve the most dominant components of variance within the text prototype manifold, assuming these capture the most salient semantic information. Based on our experiments $k_{98\%}$ is typically less than $k'$.

\subsection{Gavish-Donoho Optimal Hard Thresholding}
\label{ssec:gavish_donoho}
The Gavish-Donoho method \cite{gavish2017optimal} offers a theoretically grounded approach for selecting an optimal number of singular values $k_t$, to retain, particularly when seeking a robust low-rank representation of the data. Developed from random matrix theory, this method computes an optimal singular value threshold $\omega^\star$. This threshold is designed to effectively separate the more dominant and structured components within the singular value spectrum from those that are less influential or exhibit characteristics similar to the singular values of a random matrix. The singular values $\sigma_i < \omega^\star$ (and their corresponding singular vectors) are consequently excluded, leading to the determination of the rank $k_t$. The specific threshold value depends on the aspect ratio of the matrix undergoing SVD and can be established using the properties of the singular value spectrum itself (e.g., via the median singular value), thus providing a data-driven cutoff without requiring an explicit "noise" model. We apply this method to determine $k_t$ for our text prototype matrix, in order to identify a subset of singular vectors that is the basis of adaptation.

\subsection{Performance Impact of Singular Vector Selection}
\label{ssec:svd_selection_results}
To demonstrate the efficacy of these selection strategies, we evaluate our TTA method on ImageNet-A dataset. Table~\ref{tab:svd_selection_comparison} presents the top-1 accuracy, comparing the zero-shot CLIP baseline \cite{radford2021learning} with our TTA approach when using singular vectors selected by the 98\% energy criterion ($k_{98\%}$) versus the Gavish-Donoho method ($k_{\text{t}}$).

\begin{table*}[htbp]
  \centering
  \caption{Impact of singular vector selection on Test-Time Adaptation (TTA) performance. Average Top-1 accuracy (\%) over 3 random seeds is reported. Both selection methods significantly improve over the zero-shot baseline, with Gavish-Donoho (GD) offering a slight further advantage.}
  \label{tab:svd_selection_comparison}
  \begin{tabular}{@{}lccc@{}}
    \toprule
    Dataset & Zero-Shot & STS (Ours) & STS (Ours) \\
            & (ViT-B/16)     & ($k_{98\%}$ SVs) & ($k_{\text{t}}$ SVs) \\
    \midrule
    ImageNet-A \cite{hendrycks2021natural} & 47.87 & 61.09 (+13.22) & \textbf{61.23 (+13.36)} \\
    \bottomrule
  \end{tabular}
\end{table*}

The results in Table~\ref{tab:svd_selection_comparison} clearly indicate that employing a principled selection of singular vectors substantially boosts the performance of our TTA method compared to the zero-shot baseline. Both the 98\% energy criterion and the Gavish-Donoho threshold lead to significant improvements. Notably, the Gavish-Donoho method consistently achieves slightly better performance, suggesting its effectiveness in identifying an optimal rank for the text prototype subspace used in our adaptation process. This underscores the importance of focusing the adaptation on the most informative and semantic dimensions derived from the text prototypes.

\section{Additional Implementation Details}
\label{datasets}
\subsection{Dataset Details}
In Table \ref{tab:datasets_stats}, we present the detailed statistics of each dataset we used in our experiments, including the number of classes, the sizes of training, validation and testing sets, and their original tasks.
\begin{table*}[htbp]
\centering
\caption{Detailed statistics of datasets used in experiments. Note that the 4 ImageNet variant datasets are designed for evaluation and only contain the test sets.}
\label{tab:datasets_stats}
\resizebox{\textwidth}{!}{  
\begin{tabular}{@{}lrrrrl@{}}
\toprule
Dataset & Classes & Training & Validation & Testing & Task \\
\midrule
ImageNet \cite{deng2009imagenet} & 1,000 & 1.28M & - & 50,000 & Object recognition \\
ImageNet-A \cite{hendrycks2021natural}  & 200 & - & - & 7,500 & Robustness of adversarial attack \\
ImageNet-V2 \cite{recht2019imagenet}  & 1,000 & - & - & 10,000 & Robustness of collocation \\
ImageNet-R \cite{hendrycks2021many} & 200 & - & - & 30,000 & Robustness of multi-domains \\
ImageNet-Sketch \cite{wang2019learning}  & 1,000 & - & - & 50,889 & Robustness of sketch domain \\
\midrule
Caltech101 \cite{fei2004learning} & 100 & 4,128 & 1,649 & 2,465 & Object recognition \\
DTD \cite{cimpoi2014describing} & 47 & 2,820 & 1,128 & 1,692 & Texture recognition \\
EuroSAT \cite{helber2019eurosat} & 10 & 13,500 & 5,400 & 8,100 & Satellite image recognition \\
FGVCAircraft \cite{maji2013fine} & 100 & 3,334 & 3,333 & 3,333 & Fine-grained aircraft recognition \\
Flowers102 \cite{nilsback2008automated}  & 102 & 4,093 & 1,633 & 2,463 & Fine-grained flowers recognition \\
Food101 \cite{bossard2014food} & 101 & 50,500 & 20,200 & 30,300 & Fine-grained food recognition \\
OxfordPets \cite{parkhi2012cats} & 37 & 2,944 & 736 & 3,669 & Fine-grained pets recognition \\
StanfordCars \cite{krause20133d} & 196 & 6,509 & 1,635 & 8,041 & Fine-grained car recognition \\
SUN397 \cite{xiao2010sun} & 397 & 15,880 & 3,970 & 19,850 & Scene recognition \\
UCF101  \cite{soomro2012ucf101}  & 101 & 7,639 & 1,898 & 3,783 & Action recognition \\
\bottomrule
\end{tabular}
}
\end{table*}

In Table \ref{tab:dataset_prompts_ensemble_separated}, we detail the specific hand-crafted prompts utilized in our experiments.

\begin{table*}[htbp]
\centering
\caption{Datasets with associated textual prompts. The first prompt is applied generally, while the subsequent \textit{generic} prompts (indicated by the brace) are collectively used as an ensemble for each dataset. These 7 generic templates are highlighted in the official CLIP repository \cite{clip2021github}.}
\label{tab:dataset_prompts_ensemble_separated}
\resizebox{0.8\textwidth}{!}{  
\begin{tabular}{@{}lc@{}}
\toprule
Dataset & Prompts \\
\midrule
ImageNet  \cite{deng2009imagenet}      & \multirow{15}{*}{\begin{tikzpicture} 
    \node (p_outer) [align=left, anchor=west] at (0,0) 
                   {``a photo of a \texttt{\{CLASS\}}.''};

    \matrix (ensemble_matrix) [matrix of nodes,
                                 nodes={align=left, anchor=west}, 
                                 row sep=0.05cm, 
                                 anchor=north west] 
                                 at ([yshift=-0.25cm]p_outer.south west) 
    { 
      ``a bad photo of the \texttt{\{CLASS\}}.'' \\
      ``a \texttt{\{CLASS\}} in a video game.'' \\
      ``a origami \texttt{\{CLASS\}}.'' \\
      ``a photo of the small \texttt{\{CLASS\}}.'' \\
      ``art of the \texttt{\{CLASS\}}.'' \\
      ``a photo of the large \texttt{\{CLASS\}}.'' \\
      ``itap of a \texttt{\{CLASS\}}.'' \\
    };

    \coordinate (BraceDrawStart) at ([xshift=0.1cm]ensemble_matrix.north east); 
    \coordinate (BraceDrawEnd)   at ([xshift=0.1cm]ensemble_matrix.south east);

    \draw [decoration={brace, amplitude=5pt}, decorate] 
          (BraceDrawStart) -- (BraceDrawEnd);

    \coordinate (BraceMid) at ($ (BraceDrawStart) !0.5! (BraceDrawEnd) $);

    \coordinate (ArrowStart) at ([xshift=6pt]BraceMid); 
    \node (EnsembleText) [right=0.6cm of BraceMid] {Ensemble}; 

    \draw [-{Stealth[length=2.5mm, width=2.5mm]}, thick] (ArrowStart) -- (EnsembleText.west);
\end{tikzpicture}} \\ 
ImageNet-V2 \cite{recht2019imagenet}    & \\
ImageNet-Sketch  \cite{wang2019learning}& \\
ImageNet-A  \cite{hendrycks2021natural}   & \\
ImageNet-R   \cite{hendrycks2021many}  & \\
Caltech101   \cite{fei2004learning}   & \\
DTD      \cite{cimpoi2014describing}       & \\
EuroSAT   \cite{helber2019eurosat}      & \\
FGVCAircraft  \cite{maji2013fine}  & \\
Flowers102  \cite{nilsback2008automated}    & \\
Food101     \cite{bossard2014food}    & \\
OxfordPets  \cite{parkhi2012cats}    & \\
StanfordCars \cite{krause20133d}   & \\
SUN397   \cite{xiao2010sun}       & \\
UCF101   \cite{soomro2012ucf101}       & \\
\bottomrule
\end{tabular}
}
\end{table*}

\section{License Information}
\label{licenses}
\paragraph{Datasets.} We list the known license information for the datasets below: 
\begin{itemize}
    \item CC BY-SA 4.0 License: OxfordPets \cite{parkhi2012cats}.
    \item MIT License: ImageNet-A \cite{hendrycks2021natural}, ImageNet-V2 \cite{recht2019imagenet}, ImageNet-R \cite{hendrycks2021many}, and ImageNet-Sketch \cite{wang2019learning}. 
    \item Research Purposes only (term of access): ImageNet \cite{deng2009imagenet}, DTD \cite{cimpoi2014describing}, StanfordCars \cite{krause20133d}, SUN397 \cite{xiao2010sun}, FGVCAircraft \cite{maji2013fine}. 
\end{itemize}
\paragraph{Source Code.} In this work, we also use some code implementations from existing baseline methods to report their results: CLIP \cite{radford2021learning} , CoOp \cite{zhou2022learning}, MaPLe \cite{khattak2023maple}, TPT \cite{shu2022test}. The source code used in this paper for these methods is available under the MIT License.

\end{document}